\documentclass[runningheads]{llncs}
\usepackage{graphicx}
\usepackage{amsmath,amssymb} 
\usepackage{color}

\usepackage[utf8]{inputenc} 
\usepackage[T1]{fontenc}    
\usepackage{hyperref}       
\usepackage{url}            
\usepackage{booktabs}       
\usepackage{amsfonts}       
\usepackage{nicefrac}       
\usepackage{microtype}      
\usepackage{subcaption}
\usepackage{multirow}
\usepackage{makecell}
\usepackage{cite}

\newcolumntype{?}[1]{!{\vrule width #1}}

\newcommand{\hbline}{\Xhline{2.5\arrayrulewidth}}

\newcommand\nocell[1]{\multicolumn{#1}{c?{1.5pt}}{}}

\setcellgapes{5pt}
\renewcommand{\tabcolsep}{5mm}

\usepackage{color}

\begin{document}
\title{CapsuleGAN: Generative Adversarial \\Capsule Network}

\titlerunning{CapsuleGAN}
%

\author{Ayush Jaiswal, Wael AbdAlmageed, Yue Wu, Premkumar Natarajan}

%
\authorrunning{Jaiswal et al.}

\institute{USC Information Sciences Institute\\
Marina del Rey, CA, USA\\
\email{\{ajaiswal, wamageed, yue\_wu, pnataraj\}@isi.edu}}

\maketitle              
\begin{abstract}
We present Generative Adversarial Capsule Network (CapsuleGAN), a framework that uses capsule networks (CapsNets) instead of the standard convolutional neural networks (CNNs) as discriminators within the generative adversarial network (GAN) setting, while modeling image data. We provide guidelines for designing CapsNet discriminators and the updated GAN objective function, which incorporates the CapsNet margin loss, for training CapsuleGAN models. We show that CapsuleGAN outperforms convolutional-GAN at modeling image data distribution on MNIST and CIFAR-10 datasets, evaluated on the generative adversarial metric and at semi-supervised image classification.
\end{abstract}
%
%
%

\section{Introduction}
\label{sec:introduction}
Generative modeling of data is a challenging machine learning problem that has garnered tremendous interest recently, partly due to the invention of generative adversarial networks (GANs)~\cite{bib:gan} and its several sophisticated variants~\footnote{https://github.com/hindupuravinash/the-gan-zoo}. A GAN model is typically composed of two neural networks; (1) a generator that attempts to transform samples drawn from a  prior distribution to samples from a complex data distribution with much higher dimensionality, and (2) a discriminator that decides whether the given sample is real or from the generator's distribution. The two components are trained by playing an adversarial game. GANs have shown great promise in modeling highly complex distributions underlying real world data, especially images. However, they are notorious for being difficult to train and have problems with stability, vanishing gradients, mode collapse and inadequate mode coverage~\cite{bib:dcgan, bib:igan, bib:gman}. Consequently, there has been a large amount of work towards improving GANs by using better objective functions~\cite{bib:wgan, bib:iwgan, bib:began}, sophisticated training strategies~\cite{bib:igan}, using structural hyperparameters~\cite{bib:dcgan, bib:acgan} and adopting empirically successful tricks~\footnote{https://github.com/soumith/ganhacks}.

Radford et al.~\cite{bib:dcgan} provide a set of architectural guidelines, formulating a class of convolutional neural networks (CNNs) that have since been extensively used to create GANs (referred to as Deep Convolutional GANs or DCGANs) for modeling image data and other related applications~\cite{bib:im_synth, bib:im2im_tr}. More recently, however, Sabour et al.~\cite{bib:capsnet} introduced capsule networks (CapsNets) as a powerful alternative to CNNs, which learn a more \emph{equivariant} representation of images that is more robust to changes in pose and spatial relationships of parts of objects in images~\cite{bib:transauto} (information that CNNs lose during training, by design). Inspired by the working mechanism of optic neurons in the human visual system, capsules were first introduced by Hinton et al.~\cite{bib:transauto} as locally invariant groups of neurons that learn to recognize visual entities and output activation vectors that represent both the presence of those entities and their properties relevant to the visual task (such as object classification). The training algorithm of CapsNets involves a routing mechanism between capsules in successive layers of the network that imitates hierarchical communication of information across neurons in human brains that are responsible for visual perception and understanding.

The initial intuition behind the design of deep neural networks was to  imitate human brains for modeling hierarchical recognition of features, starting from low-level attributes and progressing towards complex entities. CapsNets capture this intuition more effectively than CNNs because they have the aforementioned in-built explicit mechanism that models it. CapsNets have been shown to outperform CNNs on MNIST digit classification and segmentation of overlapping digits~\cite{bib:capsnet}. This motivates the question whether GANs can be designed using CapsNets (instead of CNNs) to improve their performance.

We propose Generative Adversarial Capsule Network (CapsuleGAN) as a framework that incorporates capsules within the GAN framework. In particular, CapsNets are used as discriminators in our framework as opposed to the conventionally used CNNs. We show that CapsuleGANs perform better than CNN-based GANs at modeling the underlying distribution of MNIST~\cite{bib:mnist} and CIFAR-10~\cite{bib:cifar} datasets both qualitatively and quantitatively using the generative adversarial metric (GAM)~\cite{bib:gam} and at semi-supervised classification using unlabeled GAN-generated images with a small number of labeled real images.

The rest of the paper is organized as follows. Section~\ref{sec:related_work} discusses related work. In Section~\ref{sec:prelim} we provide a brief introduction to GANs and CapsNets. Section~\ref{sec:capsgan} describes our CapsuleGAN framework along with implementation guidelines. Qualitative and quantitative analyses of our model are presented in Section~\ref{sec:evaluation}. Section~\ref{sec:conclusion} concludes the paper and provides directions for future research.


\section{Related Work}
\label{sec:related_work}

GANs were originally implemented as feedforward multi-layer perceptrons, which did not perform well on generating complex images like those in the CIFAR-10 dataset~\cite{bib:cifar}. They suffered from mode collapse and were highly unstable to train~\cite{bib:dcgan, bib:igan}. In an attempt to solve these problems, Radford et al.~\cite{bib:dcgan} presented a set of guidelines to design GANs as a class of CNNs, giving rise to DCGANs, which have since been a dominant approach to GAN network architecture design. Im et al.~\cite{bib:gran} later proposed the use of Recurrent Neural Networks instead of CNNs as generators for GANs, creating a new class of GANs referred to as Generative Recurrent Adversarial Networks or GRANs. On a related note, Odena et al.~\cite{bib:acgan} proposed an architectural change to GANs in the form of a discriminator that also acts as a classifier for class-conditional image generation. This approach for designing discriminators has been a popular choice for conditional GANs~\cite{bib:cgan} recently. Our work is similar in line with~\cite{bib:acgan} in the sense that we propose an architectural change to discriminators. We propose to transition from designing GAN discriminators as CNNs to formulating them as CapsNets, creating a new class of GANs called CapsuleGANs. This idea can be extended to encoder-based GANs like BiGAN~\cite{bib:bigan} where the encoder can be modeled as a CapsNet also.


\section{Preliminaries}
\label{sec:prelim}

\subsection{Generative Adversarial Networks}
Goodfellow et al.~\cite{bib:gan} introduced GANs as a framework for generative modeling of data through learning a transformation from points belonging to a simple prior distribution ($\mathbf{z} \sim p_z$) to those from the data distribution ($\mathbf{x} \sim p_{data}$). The framework is composed of two models that play an adversarial game: a generator and a discriminator. While the generator attempts to learn the aforementioned transformation $G(\mathbf{z})$, the discriminator acts as a critic $D(\cdot)$ determining whether the sample provided to it is from the generator's output distribution ($G(\mathbf{z}) \sim p_G$) or from the data distribution ($\mathbf{x} \sim p_{data}$), thus giving a scalar output ($y \in \{0, 1\}$). The goal of the generator is to fool the discriminator by generating samples that resemble those from the real data while that of the discriminator is to accurately distinguish between real and generated data. The two models, typically designed as neural networks, play an adversarial game with the objective as shown in Equation~\ref{eq:gan}.

\begin{align}
\min_{G} \max_{D} V(D, G) = \mathbb{E}_{\mathbf{x} \sim p_{data}(\mathbf{x})} \left [ \log D(\mathbf{x}) \right ] + \ \ \mathbb{E}_{\mathbf{z} \sim p_{z}(\mathbf{z})} \left [ \log (1 - D(G(\mathbf{z}))) \right ]
\label{eq:gan}
\end{align}

\subsection{Capsule Networks}

The concept of capsules was first introduced by Hinton et al.~\cite{bib:transauto} as a method for learning robust unsupervised representation of images. Capsules are locally invariant groups of neurons that learn to recognize the presence of visual entities and encode their properties into vector outputs, with the vector length (limited to being between zero and one) representing the presence of the entity. For example, each capsule can learn to identify certain objects or object-parts in images. Within the framework of neural networks, several capsules can be grouped together to form a capsule-layer where each unit produces a vector output instead of a (conventional) scalar activation.

Sabour et al.~\cite{bib:capsnet} introduced a routing-by-agreement mechanism for the interaction of capsules within deep neural networks with several capsule-layers, which works by pairwise determination of the passage of information between capsules in successive layers. For each capsule $h_i^{(l)}$ in layer $l$ and each capsule $h_j^{(l + 1)}$ in the layer above, a coupling coefficient $c_{ij}$ is adjusted iteratively based on the agreement (cosine similarity) between $h_i$'s prediction of the output of $h_j$ and its actual output given the product of $c_{ij}$ and $h_i$'s activation. Thus, the coupling coefficients inherently decide how information flows between pairs of capsules. For a classification task involving $K$ classes, the final layer of the CapsNet can be designed to have $K$ capsules, each representing one class. Since the length of a capsule's vector output represents the presence of a visual entity, the length of each capsule in the final layer ($\left\lVert \mathbf{v_k} \right\rVert$) can then be viewed as the probability of the image belonging to a particular class ($k$). The authors introduce a margin loss $L_M$ for training CapsNets for multi-class classification, as show in Equation~\ref{eq:capsnet}:

\begin{align}
\label{eq:capsnet}
L_M = \sum_{k=1}^{K} T_k \max (0, m^+ - \left\lVert \mathbf{v_k} \right\rVert)^2 + \lambda (1 - T_k) \max(0, \left\lVert \mathbf{v_k} \right\rVert - m^-)^2
\end{align}

where $T_k$ represents target labels, $m^+ = 0.9$, $m^- = 0.1$ and $\lambda = 0.5$, a down-weighting factor for preventing initial learning from shrinking the lengths of the capsule outputs in the final layer. The authors also add regularization to the network in the form of a weighted image reconstruction loss, where the vector outputs $\mathbf{v_k}$ of the final layer are presented as inputs to the reconstruction network.


\section{Generative Adversarial Capsule Networks}
\label{sec:capsgan}
GANs have been mostly used for modeling the distribution of image data and associated attributes, as well as for other image-based applications like image-to-image translation~\cite{bib:im2im_tr} and image synthesis from textual descriptions~\cite{bib:im_synth}. The generator and the discriminator have conventionally been modeled as deep CNNs following the DCGAN guidelines~\cite{bib:dcgan}. We follow this convention in designing the CapsuleGAN generator as a deep CNN. However, motivated by the stronger intuition behind and the superior performance of CapsNets with respect to CNNs~\cite{bib:capsnet}, we design the proposed CapsuleGAN framework to incorporate capsule-layers instead of convolutional layers in the GAN discriminator, which fundamentally performs a two-class classification task.

The CapsuleGAN discriminator is similar in architecture to the CapsNet model presented in~\cite{bib:capsnet}. CapsNets, in general, have a large number of parameters because, firstly, each capsule produces a vector output instead of a single scalar and, secondly, each capsule has \emph{additional} parameters associated with all the capsules in the layer above it that are used for making predictions about their outputs. However, it is necessary to keep the number of parameters in the CapsuleGAN discriminator low due to two reasons: (1) CapsNets are very powerful models and can easily start harshly penalizing the generator early on in the training process, which will cause the generator to either fail completely or suffer from mode collapse, and (2) current implementations of the dynamic routing algorithm are slow to run. It is important to note that first reason for keeping the number of parameters of the CapsNet low falls in line with the popular design of convolutional discriminators as relatively shallow neural networks with low numbers of relatively large-sized filters in their convolutional layers.

The final layer of the CapsuleGAN discriminator contains a single capsule, the length of which represents the probability whether the discriminator's input is a real or a generated image. We use  margin loss $L_M$ instead of the conventional binary cross-entropy loss for training our CapsuleGAN model because $L_M$ works better for training CapsNets. Therefore, the objective of CapsuleGAN can be formulated as shown in Equation~\ref{eq:capsgan}.

\begin{align}
&\min_{G} \max_{D} V(D, G) \nonumber \\
&= \mathbb{E}_{\mathbf{x} \sim p_{data}(\mathbf{x})} \left [ -L_M(D(\mathbf{x}), \mathbf{T}=\mathbf{1}) \right ] + \ \ \mathbb{E}_{\mathbf{z} \sim p_{z}(\mathbf{z})} \left [ -L_M(D(G(\mathbf{z})), \mathbf{T}=\mathbf{0}) \right ]
\label{eq:capsgan}
\end{align}

In practice, we train the generator to minimize $L_M(D(G(\mathbf{z})), \mathbf{T=1})$ instead of minimizing $-L_M(D(G(\mathbf{z})), \mathbf{T}=\mathbf{0})$. This essentially eliminates the down-weighting factor $\lambda$ in $L_M$ when training the generator, which does not contain any capsules.


\section{Experimental Evaluation}
\label{sec:evaluation}

We evaluate the performance of CapsuleGANs at randomly generating images through a series of experiments as described below, in which we compare CapsuleGANs with convolutional GANs both qualitatively and quantitatively. We implement both the GAN models with the same architecture for their generators. Both convolutional GAN and the proposed CapsuleGAN models are implemented using the publicly available \texttt{keras-adversarial}~\footnote{https://github.com/bstriner/keras-adversarial} and \texttt{CapsNet-Keras}~\footnote{https://github.com/XifengGuo/CapsNet-Keras} packages.

\subsection{Data}

We provide results of our experiments on MNIST and CIFAR-10 datasets. The MNIST dataset consists of $28 \times 28$ sized grayscale images of handwritten digits. The CIFAR-10 dataset contains $32 \times 32$ color images grouped into ten classes: airplane, automobile, bird, cat, deer, dog, frog, horse, ship and truck.

\begin{figure}
\centering
\begin{subfigure}{.49\textwidth}
\centering
\includegraphics[width=0.99\textwidth]{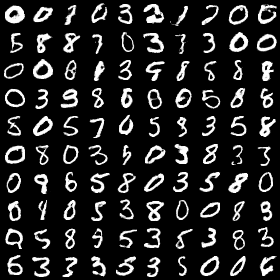}
\caption{Convolutional GAN}
\label{fig:gan_mnist_randomly_generated}
\end{subfigure}
\begin{subfigure}{.49\textwidth}
\centering
\includegraphics[width=0.99\textwidth]{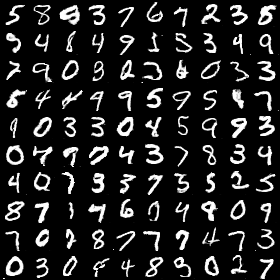}
\caption{CapsuleGAN}
\label{fig:capsgan_mnist_randomly_generated}
\end{subfigure}
\label{fig:mnist_randomly_generated}
\caption{Randomly generated MNIST images}
\end{figure}

\subsection{Visual Quality of Randomly Generated Images}

\begin{figure}
\centering
\begin{subfigure}{.49\textwidth}
\centering
\includegraphics[width=0.99\textwidth]{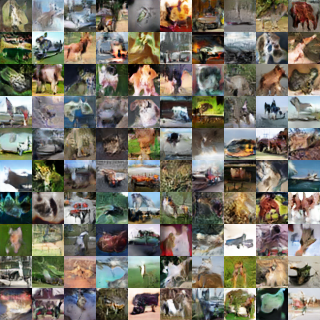}
\caption{Convolutional GAN}
\label{fig:gan_cifar10_randomly_generated}
\end{subfigure}
\begin{subfigure}{.49\textwidth}
\centering
\includegraphics[width=0.99\textwidth]{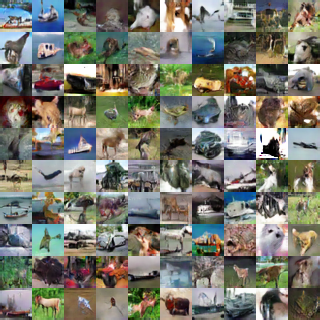}
\caption{CapsuleGAN}
\label{fig:capsgan_cifar10_randomly_generated}
\end{subfigure}
\label{fig:cifar10_randomly_generated}
\caption{Randomly generated CIFAR-10 images}
\end{figure}

We qualitatively compare images generated randomly using both GAN and CapsuleGAN. Figures~\ref{fig:gan_mnist_randomly_generated}~and~\ref{fig:capsgan_mnist_randomly_generated} show images generated using the standard convolutional-GAN and CapsuleGAN, respectively, on the MNIST dataset. Qualitatively, both CapsuleGAN and the standard convolutional-GAN produce crisp images of similar quality, that sometimes do not resemble any digit. However, the image-grid generated using GAN seems to have less diversity in terms of generated classes of digits. Figures~\ref{fig:gan_cifar10_randomly_generated}~and~\ref{fig:capsgan_cifar10_randomly_generated} show the results of this experiment on the CIFAR-10 dataset. Both the models produce diverse sets of images but images generated using CapsuleGAN look cleaner and crisper than those generated using convolutional-GAN. We provide results of our quantitative evaluation in the following subsections for deeper analyses of the image generation performance.

\subsection{Generative Adversarial Metric}

Im et al.~\cite{bib:gam} introduced the generative adversarial metric (GAM) as a pairwise comparison metric between GAN models by pitting each generator against the opponent's discriminator, i.e., given two GAN models $M_1 = (G_1, D_1)$ and $M_2 = (G_2, D_2)$, $G_1$ engages in a battle against $D_2$ while $G_2$ against $D_1$. The ratios of their classification errors on real test dataset and on generated samples are then calculated as $r_{test}$ and $r_{samples}$. Following their implementation~\footnote{https://github.com/jiwoongim/GRAN/battle.py}, in practice, the ratios of classification accuracies are calculated instead of errors to avoid numerical problems, as shown in Equations \ref{eq:gam1} and \ref{eq:gam2}

\begin{equation}\label{eq:gam1}
r_{samples} = \frac{A(D_{GAN}(G_{CapsuleGAN}(\mathbf{z})))}{A(D_{CapsuleGAN}(G_{GAN}(\mathbf{z})))}
\end{equation}

\begin{equation}\label{eq:gam2}
r_{test} = \frac{A(D_{GAN}(\mathbf{x_{test}}))}{A(D_{CapsuleGAN}(\mathbf{x_{test}}))}
\end{equation}

Therefore, for CapsuleGAN to win against GAN, both $r_{samples} < 1$ and $r_{test} \simeq 1$ must be satisfied. In our experiments, we achieve $r_{samples} = 0.79$ and $r_{test} = 1$ on the MNIST dataset and $r_{samples} = 1.0$ and $r_{test} = 0.72$ on the CIFAR-10 dataset. Thus, on this metric, CapsuleGAN performs better than convolutional GAN on the MNIST dataset but the two models tie on the CIFAR-10 dataset.

\makegapedcells
\begin{table}
\centering
\caption{Results of semi-supervised classification - MNIST}
\label{tab:mnist_semi_supervised}
\begin{tabular}{ c ?{1.5pt} c | c | c }
  \hbline
  \textbf{Model} & \multicolumn{3}{c}{\textbf{Error Rate}} \\
  \hbline
  \nocell{1} & n = 100 & n = 1,000 & n = 10,000 \\ \cline{2-4}
  Convolutional GAN & 0.2900 & 0.1539 & 0.0702 \\
  CapsuleGAN & \textbf{0.2724} & \textbf{0.1142} & \textbf{0.0531} \\
  \hbline
\end{tabular}
\end{table}

\makegapedcells
\begin{table}
\centering
\caption{Results of semi-supervised classification - CIFAR-10}
\label{tab:cifar10_semi_supervised}
\begin{tabular}{ c ?{1.5pt} c | c | c }
  \hbline
  \textbf{Model} & \multicolumn{3}{c}{\textbf{Error Rate}} \\
  \hbline
  \nocell{1} & n = 100 & n = 1,000 & n = 10,000 \\ \cline{2-4}
  Convolutional GAN & 0.8305 & 0.7587 & 0.7209 \\
  CapsuleGAN & \textbf{0.7983} & \textbf{0.7496} & \textbf{0.7102} \\
  \hbline
\end{tabular}
\end{table}

\subsection{Semi-supervised Classification}

We evaluate the performance of the convolutional GAN and the proposed CapsuleGAN on semi-supervised classification. In this experiment, we randomly generate $50,000$ images using both GAN and CapsuleGAN. We use the Label Spreading algorithm~\cite{bib:label_spreading} with the generated images as the unlabeled examples and $n$ real labeled samples, with $n \in \{100, 1000, 10000\}$. We use the \texttt{scikit-learn}~\footnote{http://scikit-learn.org/} package for these experiments. Table~\ref{tab:mnist_semi_supervised} shows the results of our experiments on MNIST while Table~\ref{tab:cifar10_semi_supervised} shows those on CIFAR-10. The error rates are high in most experimental settings because we provide raw pixel values as features to the classification algorithm. However, this allows us to more objectively compare the two models without being biased by feature extraction methods. The results show that the proposed CapsuleGAN consistently outperforms convolutional GAN for all the tested values of $n$ with a margin of $1.7-3.97$ percentage points for MNIST and $0.91-3.22$ percentage points for CIFAR-10. Thus, CapsuleGAN generates images that are more similar to real images and more diverse than those generated using convolutional GAN, leading to better semi-supervised classification performance on the test dataset.


\section{Discussion and Future Work}
\label{sec:conclusion}

Generative adversarial networks are extremely powerful tools for generative modeling of complex data distributions. Research is being actively conducted towards further improving them as well as making their training easier and more stable. Motivated by the success of CapsNets over CNNs at image-based inference tasks, we presented the generative adversarial capsule network (CapsuleGAN), a GAN variant that incorporates CapsNets instead of CNNs as discriminators when modeling image data. We presented guidelines for designing CapsuleGANs as well as an updated objective function for training CapsuleGANs. We showed that CapsuleGANs outperform convolutional-GANs on the generative adversarial metric and at semi-supervised classification with a large number of unlabeled generated images and a small number of real labeled ones, on  MNIST and CIFAR-10 datasets. This indicates that CapsNets should be considered as potential alternatives to CNNs for designing discriminators and other inference modules in future GAN models.

We plan to conduct theoretical analysis of the use of margin loss within the GAN objective. We purposefully did not incorporate many GAN training tricks to fairly evaluate our contributions. The results presented in this paper motivate the use of CapsNets as opposed to CNNs for encoders in GAN variants like BiCoGAN~\cite{bib:bcgan}. We see this as an important direction for future research.


\subsection*{Acknowledgements}
This work is based on research sponsored by the Defense Advanced Research Projects Agency under agreement number FA8750-16-2-0204. The U.S. Government is authorized to reproduce and distribute reprints for governmental purposes  notwithstanding any copyright notation thereon. The views and conclusions contained herein are those of the authors and should not be interpreted as necessarily representing the official policies or endorsements, either expressed or implied, of the Defense Advanced Research Projects Agency or the U.S. Government.

%
%
%
\bibliographystyle{splncs04}
\bibliography{egbib}

\end{document}